# OPTIMIZATION OF SLIDING CONTROL PARAMETERS FOR A 3-DOF ROBOT ARM USING GENETIC ALGORITHM (GA)

TỐI ƯU HÓA THAM SỐ BỘ ĐIỀU KHIỂN TRƯỢT CHO TAY MÁY ROBOT 3 BẬC SỬ DỤNG THUẬT TOÁN DI TRUYỀN

**Vu Ngoc Son[1,*], Pham Van Cuong[1,2], Dao Thi My Linh[2], Le Tieu Nien[1,3]**

**ABSTRACT**

This paper presents a method for optimizing the sliding mode control (SMC) parameter for a robot manipulator applying a genetic algorithm (GA). The objective of the SMC is to achieve precise and consistent tracking of the trajectory of the robot manipulator under uncertain and disturbed conditions. However, the system effectiveness and robustness depend on the choice of the SMC parameters, which is a difficult and crucial task. To solve this problem, a genetic algorithm is used to locate the optimal values of these parameters that gratify the capability criteria. The proposed method is efficient compared with the conventional SMC and Fuzzy-SMC. The simulation results show that the genetic algorithm with SMC can achieve better tracking capability and reduce the chattering effect.

***Keywords:*** *Genetic Algorithm (GA); SMC controller; robot manipulator; 3-DOF robot manipulator.*

**TÓM TẮT**

Bài báo này trình bày một phương pháp để tối ưu hóa tham số bộ điều khiển trượt (SMC) cho một tay máy robot bằng cách sử dụng thuật toán di truyền (GA). Mục tiêu của SMC là đạt được sự theo dõi chính xác và nhất quán của quỹ đạo của tay máy robot trong điều kiện không chắc chắn và có nhiễu loạn. Tuy nhiên, hiệu suất và sự ổn định của hệ thống phụ thuộc vào sự lựa chọn của các tham số SMC, điều này là một nhiệm vụ khó khăn và quan trọng. Để giải quyết vấn đề này, một thuật toán di truyền được sử dụng để tìm ra các giá trị tối ưu của các tham số này mà đáp ứng các tiêu chí hiệu suất đáp ứng bám quỹ đạo. Phương pháp được đề xuất được so sánh hiệu với SMC thông thường và Fuzzy-SMC. Kết quả mô phỏng cho thấy rằng thuật toán di truyền kết hợp với SMC có thể đạt được hiệu suất theo dõi tốt hơn và giảm hiện tượng rung lắc.

***Từ khoá:*** *Thuật toán di truyền; bộ điều khiển trượt; tay máy robot; robot 3 bậc tự do.*

[1]Faculty of Electrical Engineering, Hanoi University of Industry, Vietnam
[2]Thai Binh University, Vietnam
[3]East Asia University of Technology, Vietnam
*Email: vungocson.haui.271199@gmail.com

## 1. INTRODUCTION

The control of robotic manipulators has gained significant attention in various fields such as industry, research, and education. The ability to accurately control the motion of these manipulators is crucial for achieving desired tasks efficiently and reliably. This paper focuses on the cylindrical manipulator, a type of robot arm that has three degrees of freedom (DOF) with the position of each joint computed from a trajectory in the Cartesian space using the inverse kinematic model that represents the real manipulator [1]. The cylindrical manipulator has many applications in industry, research and education, such as assembly, welding, painting, pick-and-place, testing, teaching and learning. However, controlling the cylindrical manipulator is a difficult task because of its nonlinear and uncertain dynamics, coupled with the presence of external disturbances and measurement noises. Robotic manipulators need to operate stably and efficiently, which requires understanding their trajectory and controlling and monitoring them effectively. Many nonlinear control and motion control techniques have been developed to achieve this, such as optimal, adaptive, and robust control [2-6]. However, these methods have some drawbacks, such as sensitivity to parameter changes, chattering, tuning difficulty, or lack of comparison or validation. Adaptive fuzzy control and adaptive neural network control are two nonlinear control techniques that can handle uncertainties and disturbances by using fuzzy systems or neural networks to approximate the optimal control function, and using adaptive laws to estimate the unknown parameters [7-16]. However, these techniques also face some challenges, such as selecting and tuning the parameters of fuzzy systems or neural networks. Therefore, optimization methods are needed to enhance the capability and robustness of the nonlinear control techniques for robot manipulators. Optimization methods are mathematical tools that can find the best solution for a given problem with some constraints and criteria. Many optimization methods have been combined with nonlinear control techniques to search for the optimal control functions for uncertain robots in an automatic and efficient way [17-22]. These methods are population-based algorithms that generate multiple outcomes on every run. these algorithms are classified into five types: swarm intelligence, physical, chemical, human and evolutionary-based algorithms. These algorithms are inspired by natural phenomena and have two phases of search: exploitation and exploration.

This paper proposes a robust control method for the 3-DOF cylindrical manipulator using sliding mode control (SMC) and genetic algorithm. SMC is a nonlinear control algorithm that can handle nonlinear and uncertain systems effectively by driving the system state to a predefined surface in the state space and keeping it there, regardless of the uncertainties and disturbances. Genetic algorithm is applied to optimize the control parameters of SMC to achieve superior capability and reduce manual tuning. We compare our method with human expertise [23] and fuzzy logic [24] for cylindrical manipulators. We show simulation results to demonstrate the effectiveness of our method in terms of tracking accuracy, robustness and control effort. In [27], various intelligent control methods for robot manipulators are reviewed, such as neuro-fuzzy, fuzzy logic, neural network, genetic algorithm, particle swarm optimization, etc. Overall, we developed a robust sliding mode controller that can effectively control and predict the positions, velocities, and accelerations of the Cylindrical Manipulator. Furthermore, we established stability through Lyapunov analysis, which provides a solid theoretical foundation for ensuring the reliability and robustness of our control algorithm in real-world applications. We also presented simulation results to verify the capability and responsiveness of the controller when the sliding mode controller parameters are optimized using the genetic algorithm.

This article is organized as follows. Section 2 shows the dynamic model of the cylindrical manipulator. Section 3 presents the design of the sliding mode controller for the cylindrical manipulator. Section 4 explains the genetic algorithm to optimize the sliding mode controller parameter and the stability analysis of the controller. Section 5 shows the simulation results for the cylindrical manipulator using the proposed method and compares it with the SMC and Fuzzy-SMC. Section 6 concludes the paper and discusses the future work.

## 2. DYNAMIC OF CYLINDRICAL MANIPULATOR

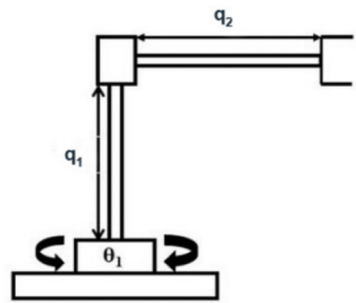

Fig. 1. The cylindrical manipulator

We consider the dynamic control equations of the Robot Manipulator. These equations capture the essence of the interaction between position, velocity, acceleration, and torque of the manipulator, enabling us to analyze and efficiently control its motion [1].

The Lagrangian formulation based on the mechanical system is defined as:

$$L(q,\dot{q}) = K(q,\dot{q}) - P(q) \quad (1)$$

the dynamics of a cylindrical manipulator with external disturbance can be expressed in the Lagrange as follows:

$$M(q)\ddot{q} + C(q,\dot{q})\dot{q} + F(q)q + G(q) = \tau - f_{ext} \quad (2)$$

where: $(q,\dot{q},\ddot{q}) \in R^{n \times 1}$ are the vectors of link position, velocity, and acceleration, respectively. $M(q) \in R^{n \times n}$ is the inertial matrix. $C(q,\dot{q}) \in R^{(n \times n)}$ is the vector of Coriolis and centripetal forces. $G(q) \in R^{n \times 1}$ expresses the gravity vector. $F(q)$ represents the vector of the frictions. $f_{ext} \in R^{n \times 1}$ is the unknown disturbances input vector. $\tau \in R^{n \times 1}$ is the link torque input vector.

## 3. DESIGN SMC CONTROLLER FOR CYLINDRICAL MANIPULATOR

Consider cylindrical manipulator dynamic system:

$$x^{(n)} = f(x) + b(x,u) \quad (3)$$

The SMC controller is designed by defining the sliding mode surface and a switching function responsible for minimizing chattering effect. Define the errors $e_1$, $e_2$ and $e_3$ such that:

$$e_1 = \theta_{1d} - \theta_1$$
$$e_2 = q_{2d} - q_2 \qquad (4)$$
$$e_3 = q_{3d} - q_3$$

The sliding mode surface is built such as:

$$S = \begin{bmatrix} s_1 \\ s_2 \\ s_3 \end{bmatrix} = \begin{bmatrix} c_1 e_1 + c_2 \dot{e}_1 \\ c_3 e_2 + c_4 \dot{e}_2 \\ c_5 e_3 + c_6 \dot{e}_3 \end{bmatrix} \qquad (5)$$

and hurwitz polynomial:

$$A(S) = 1 + \lambda_1 S + \lambda_2 S + \lambda_3 S \qquad (6)$$

we have:

$$\lim_{t \to \infty} e_i(t) = 0 \to S(0) = 0 \quad (i = 1 \div 3) \qquad (7)$$

$$\dot{s}_1 = c_1 \dot{e}_1 + c_2 \ddot{e}_1 = c_1(\dot{\theta}_{1d} - \dot{\theta}_1) + c_2 \ddot{\theta}_{1d} - c_2(f_1(x) + g_1(x,u))$$
$$= -\lambda_1 sign(s_1)$$
$$\dot{s}_2 = c_3 \dot{e}_2 + c_4 \ddot{e}_2 = c_3(\dot{q}_{2d} - \dot{q}_2) + c_4 \ddot{q}_{2d} - c_4(f_2(x) + g_2(x,u)) \qquad (8)$$
$$= -\lambda_2 sign(s_2)$$
$$\dot{s}_3 = c_5 \dot{e}_3 + c_6 \ddot{e}_3 = c_5(\dot{q}_{3d} - \dot{q}_3) + c_6 \ddot{q}_{3d} - c_6(f_3(x) + g_3(x,u))$$
$$= -\lambda_3 sign(s_3)$$

$$\begin{bmatrix} u_{dk1} \\ u_{dk2} \\ u_{dk3} \end{bmatrix} = \begin{bmatrix} (\lambda_1 sign(s_1) + \dot{e}_1 + c_2 \dot{\theta}_{1d})/c_2 \\ (\lambda_2 sign(s_2) + \dot{e}_2 + c_4 \dot{q}_{2d})/c_4 \\ (\lambda_3 sign(s_3) + \dot{e}_3 + c_6 \dot{q}_{3d})/c_6 \end{bmatrix}$$
$$+ \begin{bmatrix} c_1 & c_2 \\ c_3 & c_4 \\ c_5 & c_6 \end{bmatrix} \begin{bmatrix} \dot{\theta}_1 \\ \dot{q}_2 \\ \dot{q}_3 \end{bmatrix} - \begin{bmatrix} g_1 \\ g_2 \\ g_3 \end{bmatrix} \qquad (9)$$

Based on Lyapunov stability analysis, when the Lyapunov function is positive definite and its derivative is negative semidefinite, the control system achieves stability. Hence, to ensure stability of the overall control system, Lyapunov function is choosen:

$$V = \frac{1}{2} S^2 \qquad (10)$$

This function is a continuous and differentiable function on the state space of the sliding controller. Differentiating V along to time, we have:

$$\dot{V}(t) = S\dot{S} \qquad (11)$$

if $\dot{V}(t) < 0$ then $V \to 0$ therefore $S \to 0$ end $e \to 0$. Sliding condition is $S\dot{S} < 0$. In that case, the sliding condition ensures that the system maintains global stability, convergence, and adherence despite model inaccuracies and disturbances.

$$S\dot{S} \leq -\alpha \|S\| \leq 0; \alpha > 0; \|S\| = \sqrt{\sum_{i=1}^{n} S_i^2} \qquad (12)$$

In (9) and (12), we have: u(t) satisfy $\dot{s}_i = -\lambda_i sign(s_i)$ ; $i = 1$ to $3$. We have:

$$\dot{V}(t) = S\dot{S} = -S\lambda sign(S) = -\lambda |S| \qquad (13)$$

for all $s_i \neq 0 \to |S| > 0 \to -\lambda |S| < 0; \lambda =$ constant
$\Rightarrow \dot{V}(t) < 0; \ \forall S \neq 0$ and $S \to 0 \Rightarrow e \to 0$

## 4. GENETIC ALGORITHMS WITH SLIDING MODE CONTROLLER

The algorithm is applied to search for the best control parameter sets for the sliding mode controller, which is the genetic algorithm. Fig. 2 is the flowchart of the genetic algorithm.

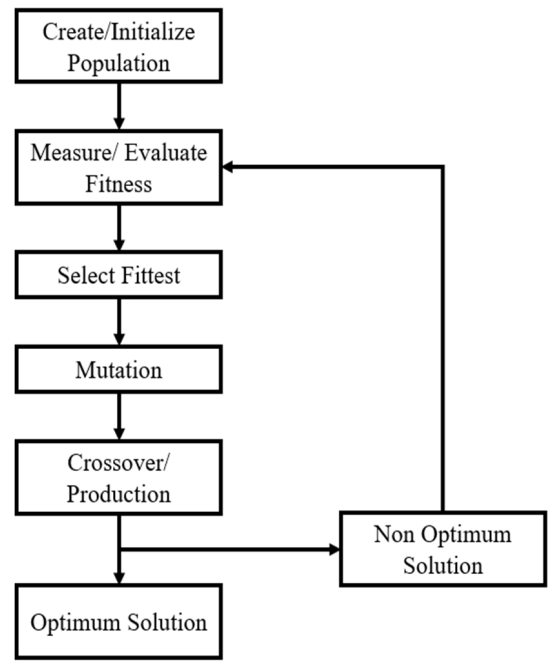

Fig. 2. The Genetic Algorithm process [28]

We developed a simulation model of a cylindrical robot manipulator based on its dynamic equations (2). The model has control signals as inputs and joint trajectory signals as outputs. We used a mathematical model derived from the dynamic properties of the vertical arm manipulator to simulate the physical system. We also applied the genetic algorithm to optimize the sliding mode controller parameters. The genetic algorithm encoded the parameters into genes and used a sequence of real numbers to represent each individual in the population.

We used the normalized integral square error (ISE) to measure the accuracy of each individual in the population. The lower the ISE, the higher the adaptability. We applied the genetic algorithm and the sliding mode controller to find the optimal parameters that minimize the ISE and the deviation from the target.

$$\text{fitness} = \int_0^\infty \left( e_1^2(t) + e_2^2(t) + e_3^2(t) \right) dt \quad (14)$$

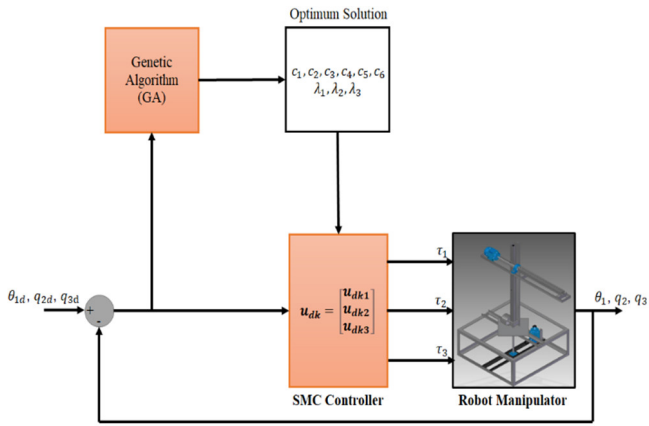

Fig. 3. Structure Genetic Algorithms with Sliding Mode Controller

## 5. SIMULATION RESULTS

In [1], equations dynamics of a cylindrical manipulator are written as the standard matrix form:

$$\begin{bmatrix} \tau_1 \\ \tau_2 \\ \tau_3 \end{bmatrix} = \begin{bmatrix} (4m_1\sin\theta_1 - 4m_2\cos\theta_1)q_3 + I_3 & 0 & (m_1+m_2)(\sin\theta_1\cos\theta_1)q_3 \\ 0 & m_3 & 0 \\ m_1\sin\theta_1\cos\theta_1 & 0 & 2(m_1\sin\theta_1 + m_2\cos\theta_1) \end{bmatrix} \times \begin{bmatrix} \ddot\theta_1 \\ \ddot q_2 \\ \ddot q_3 \end{bmatrix}$$

$$+ \begin{bmatrix} (m_1\sin\theta_1 - m_2\cos\theta_1)q_3 & 0 & -m_1\cos\theta_1 + m_2\sin\theta \\ 0 & 0 & 0 \\ 2q_3(m_1\sin\theta_1 - m_2\cos\theta_1) & 0 & 0 \end{bmatrix} \times \begin{bmatrix} \dot\theta_1^2 \\ \dot q_2^2 \\ \dot q_3^2 \end{bmatrix}$$

$$+ \begin{bmatrix} 0 & -(m_1+m_2)(\sin\theta_1\cos\theta_1)q_3 & 0 \\ 0 & 0 & 0 \\ 0 & -(m_1+m_2)(\sin\theta_1\cos\theta_1) & 0 \end{bmatrix} \times \begin{bmatrix} \dot\theta_1\dot q_2 \\ \dot\theta_1\dot q_3 \\ \dot q_2\dot q_3 \end{bmatrix} + \begin{bmatrix} 0 \\ g(m_2+m_3) \\ 0 \end{bmatrix}$$

in which $m_i$ are the mass of joint of Cylindrical Manipulator, respectively, $I_3 = 1 kg/m^2$ is the moment of inertia of joint 3, respectively, and $g = 9.8 m/m^2$ is acceleration of gravity. The values of the parameters of each joint of the manipulator are shown in Table 1.

Table 1. Values of masses (m) of each joint [1]

| Joint | m (kg) |
|---|---|
| 1 ($\theta_1$) | 36.367405 |
| 2 ($q_2$) | 12.632222 |
| 3 ($q_3$) | 23.735183 |

The position trajectories of the Cylindrical Manipulator are chosen: $\begin{bmatrix} \theta_1 & q_2 & q_3 \end{bmatrix}_0^T = \begin{bmatrix} 0.01 & 0.01 & 0.01 \end{bmatrix}^T$ and initial velocities of joints are $\begin{bmatrix} \dot\theta_1 & \dot q_2 & \dot q_3 \end{bmatrix}_0^T = \begin{bmatrix} 0 & 0 & 0 \end{bmatrix}^T$.

The simulation genetic algorithm is performed with a maximum number of generations of 1000. The number of parents used in the population is 20. The coefficient of hybridization in the population is 0.8. The coefficient of mutation in the population is 0.2. The convergence condition is fitness < 0.001. Range of values for initializing the parameters of SMC controller is 0-100.

The algorithm has converged after 134 generations and give the results in the following Table 2.

Table 2. SMC controller parameter by genetic algorithm

| Parameter $s_1$ | $c_1 = 2.1815$ | $c_2 = 0.0008$ | $\lambda_1 = 52.0574$ |
|---|---|---|---|
| Parameter $s_2$ | $c_3 = 2.2091$ | $c_4 = 0.0012$ | $\lambda_2 = 47.3860$ |
| Parameter $s_3$ | $c_5 = 1.6590$ | $c_6 = 0.0002$ | $\lambda_3 = 49.9532$ |

The desired position trajectory for the robot manipulator is chosen as follows: $\theta_{1d} = q_{2d} = q_{3d} = \sin(2\pi t + \pi/2)$

This section presents the simulation results of the Cylindrical Manipulator with the sliding mode controller with the parameters optimized by genetic algorithm in Table 2. We compare the capability and stability of the proposed sliding mode controller with two other controllers: the sliding mode controller [23] and the fuzzy-SMC controller [24]. The simulation results for the following cases are shown.

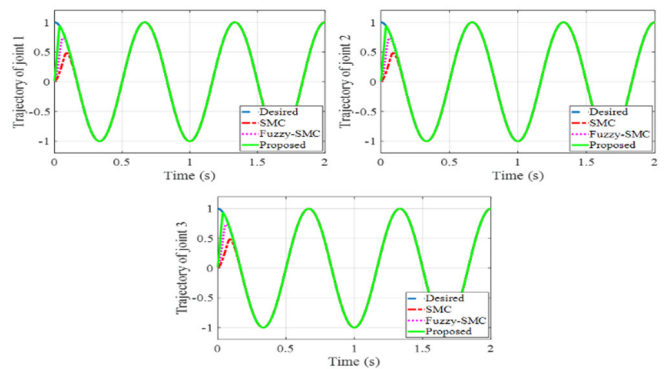

Fig. 4. Trajectory of three joint of the Cylindrical Manipulator

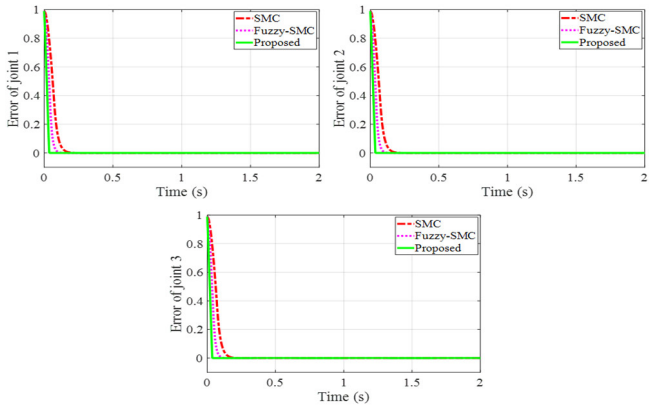

Fig. 5. Error of three joint of the Cylindrical Manipulator

The Figs. 4 and 5 show that the GA-SMC, Fuzzy-SMC, and SMC controllers all have the ability to track the desired trajectory very well, with zero overshoot. This indicates that GA-SMC has a higher capability than the other two controllers. A high-magnitude disturbance impacted to joint 3 of the Cylindrical Manipulator at 0.5s, when the robot was operating stably. We observed and compared the trajectory response results of the robot before and after the disturbance.

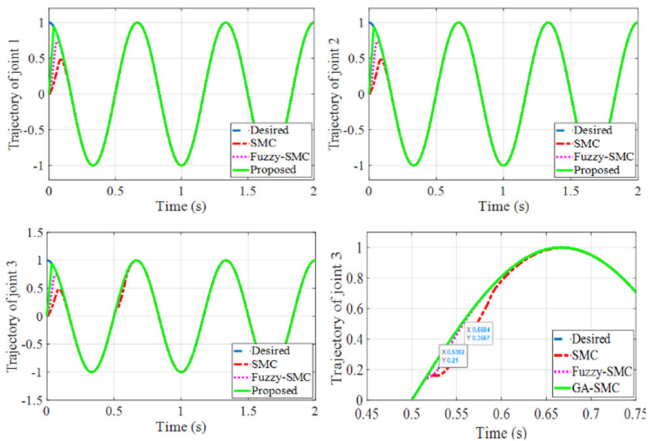

Fig. 6. Trajectory of three joint with the joint 3 disturbance

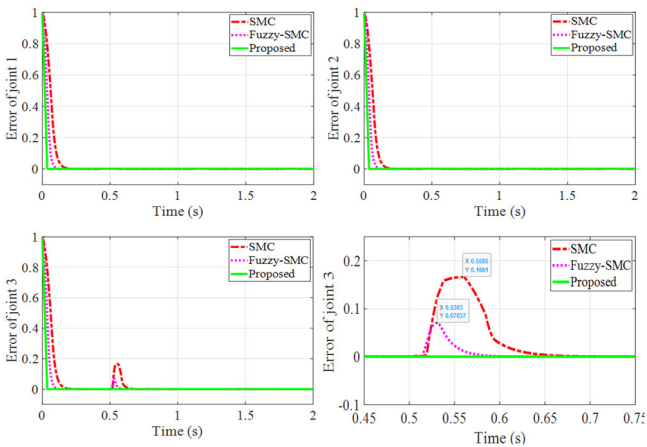

Fig. 7. Error of three joint with the joint 3 disturbance

The Figs. 6 and 7 show that the simulated controllers still have the ability to track the trajectory and response time well, as in case 1. When a large disturbance affects joint 3, SMC and Fuzzy-SMC are affected by the disturbance and lose their trajectory tracking capability. Only GA-SMC adapts strongly to the disturbance and maintains a high trajectory tracking capability. Specifically, the SMC controller achieves a maximum error (e_max) of 0.5303 units and the Fuzzy-SMC controller has a maximum error (e_max) of 0.5585 units under disturbance conditions. This demonstrates that GA-SMC has superior stability and disturbance rejection ability compared to the other two controllers.

## 6. CONCLUSION

In this work, a genetic algorithm was applyed to optimize the sliding mode control parameter of a cylindrical manipulator. The objective of the method was to enhance the tracking capability and stability of the system under uncertain and disturbed conditions. The optimal values of the sliding mode control parameters that satisfied the capability criteria were found by the genetic algorithm. The proposed method was compared with the conventional sliding mode control and fuzzy-sliding mode control in terms of tracking accuracy and chattering reduction. The results of the simulation demonstrated that the proposed method achieved better capability than the other methods. The proposed method can be extended to other types of manipulators and control systems.

**THÔNG TIN TÁC GIẢ**

**Vũ Ngọc Sơn[1], Phạm Văn Cường[1,2], Đào Thị Mỹ Linh[2], Lê Tiểu Niên[1,3]**

[1]Khoa Điện, Trường Đại học Công nghiệp Hà Nội

[2]Trường Đại học Thái Bình

[3]Trường Đại học Công nghệ Đông Á